\documentclass[10pt,journal,compsoc]{IEEEtran}

\usepackage{cite}
\usepackage{url}
\usepackage{ragged2e}
\usepackage{epsfig}
\usepackage{graphicx}
\usepackage{amsmath,amssymb} 
\usepackage{subfigure}
\usepackage{algorithm}
\usepackage{algorithmicx}
\usepackage{algpseudocode}
\usepackage{graphics}
\usepackage{threeparttable}
\usepackage{color}
\usepackage[normalem]{ulem}
\usepackage{multirow}
\usepackage{float}
\usepackage{amsfonts}
\usepackage{bm}
\usepackage{array}
\usepackage[table]{xcolor}
\usepackage{colortbl}
\usepackage{pifont}
\usepackage{diagbox}
\usepackage{rotating}
\usepackage{booktabs}
\usepackage{overpic}
\usepackage{textcomp}
\usepackage{contour}
\usepackage{enumitem}
\usepackage[colorlinks=true]{hyperref}

\newcommand{\todo}[1]{\textcolor[rgb]{0.00,0.00,0.00}{#1}}

\RequirePackage{silence}
\definecolor{bblue}{rgb}{0,150,230}
\definecolor{mygray}{gray}{.92}

\graphicspath{{./Imgs/}}
\DeclareGraphicsExtensions{.jpg,.pdf,.png}

\makeatletter
\def\ps@IEEEtitlepagestyle{%
  \def\@oddfoot{\mycopyrightnotice}%
  \def\@evenfoot{}%
}
\def\mycopyrightnotice{%
  {\hfill \scriptsize {}\hfill}

}
\makeatother

\begin{document}

\title{Cross-Modal Distillation For Widely Differing Modalities}

\author{Cairong Zhao,
        Yufeng Jin,
        Zifan Song,
        Haonan Chen,
        Duoqian Miao,
        Guosheng Hu

        \thanks{
         {Cairong Zhao, Yufeng Jin contributed equally to this work.}
         }
        \thanks{
         {Corresponding author: Guosheng Hu.}
         }

         \thanks{{Cairong Zhao, Yufeng Jin, Zifan Song, and Duoqian Miao are with the Department of Computer Science \& Technology, Tongji University, Shanghai 201804, China (e-mail: 
         zhaocairong@tongji.edu.cn,
         jinyufeng@tongji.edu.cn, 
         2111139@tongji.edu.cn,
         dqmiao@tongji.edu.cn).}}

         \thanks{{Haonan Chen is with the Alibaba Group, Hangzhou 310000, China (e-mail: haolan.chn@alibaba-inc.com).}}
         
         \thanks{{Guosheng Hu is with Oosto, Belfast BT1 2BE, UK (e-mail: huguosheng100@gmail.com).}}
}


\IEEEtitleabstractindextext{%
\begin{abstract}
\justifying
Deep learning achieved great progress recently, however, it is not easy or efficient to further improve its performance by increasing the size of the model. Multi-modal learning can mitigate this challenge by introducing richer and more discriminative information as input. To solve the problem of limited access to multi-modal data at the time of use, we conduct multi-modal learning by introducing a teacher model to transfer discriminative knowledge to a student model during training. However, this knowledge transfer via distillation is not trivial because the big domain gap between the widely differing modalities can easily lead to overfitting. In this work, we introduce a cross-modal distillation framework. Specifically, we find hard constrained loss, e.g. $l$2 loss forcing the student being exact the same as the teacher, can easily lead to overfitting in cross-modality distillation. To address this, we propose two \emph{soft} constrained knowledge distillation strategies at the feature level and classifier level respectively. In addition, we propose a quality-based adaptive weights module to weigh input samples via quantified data quality, leading to robust model training. We conducted experiments on speaker recognition and image classification tasks, and the results show that our approach is able to effectively achieve knowledge transfer between the commonly used and widely differing modalities of image, text, and speech.
\end{abstract}
}

\maketitle
\IEEEdisplaynontitleabstractindextext

\IEEEpeerreviewmaketitle

\IEEEraisesectionheading{\section{Introduction}}

\textcolor{black}{The rapid advancement of deep learning has revolutionized numerous fields by enabling the development of increasingly complex and powerful models. However, as model sizes continue to grow, the marginal benefits of scaling up models diminish, prompting researchers to explore alternative strategies for improving performance. One such strategy is multi-modal learning, which leverages the complementary strengths of multiple data modalities—such as images, speech, and text—to enhance task performance. While multi-modal learning has shown promise in various applications, its practical adoption is often hindered by the high cost and complexity of acquiring and processing multi-modal data. This limitation raises a critical question: how can we effectively utilize multi-modal data during training when only uni-modal data is available during deployment? To address this challenge, we propose a novel framework for cross-modal knowledge distillation, which enables the transfer of knowledge from a strong modality (e.g., images) to a weak modality (e.g., speech) during training, even when only the weak modality is available during inference. This approach is particularly valuable in scenarios where multi-modal data is expensive or impractical to collect, such as in smart devices without cameras or in low-resource environments. By leveraging the correlation between modalities, our framework allows the weak modality to learn discriminative features that improve its performance without relying on the strong modality at test time. Our work builds on the observation that existing knowledge distillation methods, which primarily focus on transferring knowledge between similar modalities or within a single modality, often fail to address the unique challenges of cross-modal transfer. These challenges include significant differences in data representation, modality-specific features, and varying data quality across modalities. To overcome these issues, we introduce a series of innovations that enhance the efficiency and robustness of cross-modal knowledge distillation.}

\begin{figure}[ht]
\centering
\includegraphics[width=\linewidth]{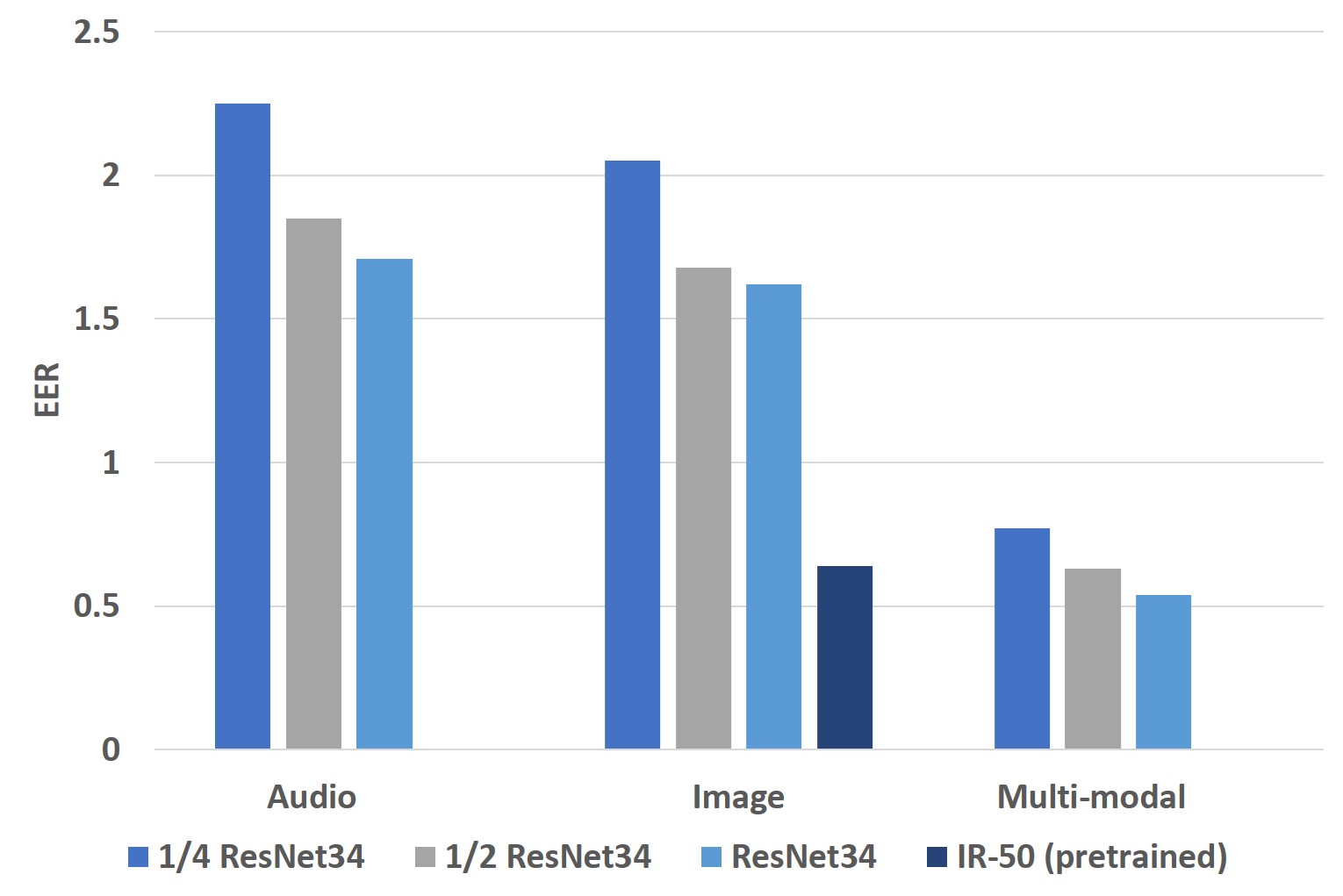}
\caption{The effect of different model sizes and different modalities on the performance of identity recognition tasks. ResNet34 with different numbers (1/4, 1/2 and full) of convolutional channels are trained on the VoxCeleb2 dataset using audio and image as inputs, respectively.}
\label{fig:motivation_exp }
\end{figure}

In order to investigate the effect of different model sizes and different modalities on the performance, we employ the identity recognition task as a case study. Specifically, we conduct training on the VoxCeleb2 dataset \cite{chung2018voxceleb2}, which is for speaker recognition with both audio and face images, using three different configurations: a quarter of convolutional channels, half channels, and the full ResNet34 model. The results are shown in Fig. \ref{fig:motivation_exp }. We can observe that the performance gains with only one modality, either audio or image, from doubling the number of parameters are progressively decreasing, indicating that it will be difficult to further improve the performance by simply increasing the model size. Nonetheless, we also find that altering the data modality (from audio to image) marginally enhances performance when the model is identical, despite this dataset not being designed for face recognition. Furthermore, different modalities can benefit from distinct training data. In the case of face images, being pre-trained on larger datasets leads to significantly increased effectiveness (IR-50 in this figure). Additionally, even when we simply calculate the average distance between the two modalities, the fusion of both is much more effective than relying on a single modality. This experiment shows that taking advantage of different modalities, multimodal learning, is an efficient path to improving the task performance.

However, the acquisition of multi-modal data poses significant challenges. In numerous practical scenarios, the collection of multi-modal data is hindered by constraints such as high costs. For example, for the aforementioned speaker recognition task, many smart speakers without cameras are unable to access multi-modal data from images and audio, even though the combination of the two modalities enables better performance. If only one modality is available at the time of use, it is difficult to efficiently improve the performance by multi-modal combination.

\begin{figure}[ht]
\centering
\includegraphics[width=0.9\linewidth]{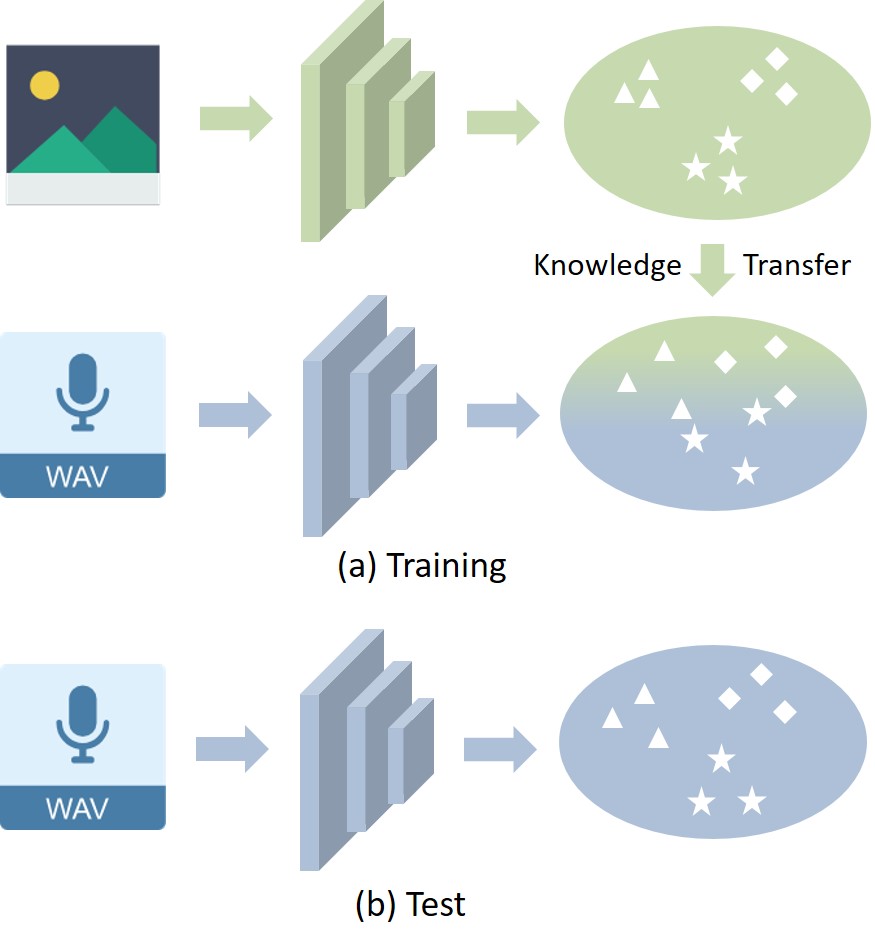}
\caption{Illustration of cross-modal knowledge transfer. The strong modality (e.g. image) transfers the discriminative knowledge to the weak modality (e.g. speech) during training. During test, the weak modality only is used. 
}
\label{fig:illustration }
\end{figure}

This knowledge transfer can be accomplished using knowledge distillation techniques \cite{hinton2015distilling}. Knowledge distillation is a widely used technique that uses a stronger teacher model to supervise a student model training, allowing knowledge transfer from the teacher to the student. Although primarily utilized for model compression, this technique can also be applied in various contexts, such as our cross-modal knowledge transfer. Existing cross-modal knowledge distillation studies basically follow the uni-modal knowledge distillation methods and are primarily conducted between similar modalities \cite{gupta2016cross, tian2019contrastive,song2023SPL,song2024alchemistcoder} (e.g. RGB images, depth maps, and textual data). 


To solve these problems, we propose our cross-modal knowledge distillation framework. Specifically, (1) In contrast to the fully fixed teacher model which is widely used by existing methods, we introduce a trainable projection head on teacher model to narrow down the gap between the teacher and the student. (2) We find the \emph{hard} alignment, e.g. using 
l
l2 loss to force the student to be the same as the teacher, is a major source of overfitting for cross-modality distillation due to the great discrepancies between two modalities.
Thus, we design two \emph{soft} constraints, feature level and classifier level, for cross-modal distillation. As for the feature level, we introduce a relaxation parameter, margin, to the existing loss functions, which do not require perfect agreement between the two modalities, allowing the student to learn modality-shared features rather than learn both modality shared and specific features. For the classifier level, we map the two modalities to the same feature space which enables them to share the classifier, implicitly making two modalities closer in this feature space rather than forcing them perfectly the same.
These two \emph{soft} alignment strategies can greatly reduce the overfitting for cross-modality distillation.
(3) We propose a quality-aware adaptive weighting module that can adaptively adjust the training objectives according to the quality of the two modal input data, avoiding the interference of low-quality data on distillation training.

{We conduct experiments in various challenging scenarios to validate our approach and successfully achieve knowledge transfer between the important and widely differing modalities of image, text, and speech. Specifically, we focus on two tasks: speaker recognition and image classification.
In speaker recognition, which usually utilizes speech to verify the identity of a speaker (during test), we aim to leverage the advancements made in face recognition, which has strong performance for identity recognition, to improve the performance of speech-based identification. 
Therefore, we use the face recognition recognition model as a teacher model to motivate the transfer of knowledge to the speaker recognition model during training.
As for image classification, this task holds great significance in the field of computer vision. Previous research \cite{radford2021learning,song2024diverse} indicates that text modal features can effectively guide image training. By constructing high-quality text data through prompts and labels, it becomes feasible to transfer knowledge from the text modality to the image modality during training.
Overall, our approach achieves promising knowledge transfer results between all these widely differing modalities, proving its broad validity.}


Our contributions can be summarized as:
\begin{itemize}
\item We find that the \emph{hard} alignment is one important source of overfitting for cross-modality distillation. Therefore, we propose two \emph{soft} alignment strategies, feature level and classifier level. 
The former introduces  a relaxation parameter, margin, in distillation loss functions, avoiding forcing the student to perfectly be the same as the teacher. The latter projects two modalities to the same feature space which share the classifier, making two modalities closer instead of forcing them to be the same. These two \emph{soft} strategies can greatly reduce the overfitting.

\item We propose a quality-based adaptive weighting module that can adaptively weight training samples based on their quality, leading to improved performance and reduced overfitting during cross-modality distillation.


\item We conduct extensive experiments on speaker recognition and image classification tasks, and our results show that our approach can effectively transfer knowledge between the commonly used and widely varying modalities of image, text, and speech, and greatly improves the uni-modality performance.

\end{itemize}

\section{Related Work}

\subsection{Knowledge Distillation}
Knowledge distillation, initially proposed by \cite{hinton2015distilling}, is a well-established technique in the field of artificial intelligence that leverages a proficient teacher model to facilitate the training of a student model. This approach utilizes the Kullback-Leibler divergence to align the logits, which are the model's outputs after the classifier. Throughout the training process, the logits are softened using higher temperature hyperparameters, enabling the extraction of more detailed negative label information. During inference, the temperature is reset to its normal value, mirroring the distillation process.

Many new knowledge distillation methods are proposed in subsequent studies. Some researchers continue to do knowledge distillation on classifiers and logits. \cite{zhao2022decoupled} proposes decoupled knowledge distillation (DKD) to decouple logits into target class and non-target class. \cite{chen2022knowledge} introduces a simple knowledge distillation (SimKD) method that reused the classifier of the teacher model to the student model, and aligned the features of the teacher model and the student model through L2 loss, achieving good alignment. Other researchers also explore feature-level knowledge distillation. \cite{zagoruyko2016paying} introduces the attention transfer (AT) method, which encourages the student model to learn the spatial attention distribution from the teacher model. Some approaches focus on channel structural features, \cite{tung2019similarity} proposes the similarity preserving (SP) method, which emphasizes the importance of maintaining consistent activation similarities between samples of both the teacher and student models. Similarly, \cite{liu2021exploring} focuses on channel correlation as the feature of knowledge transfer, proposing the inter-channel correlation for knowledge distillation (ICKD) method. \cite{passalis2018probabilistic} designs the probability knowledge transfer (PKT) method to make the eigenvector distribution of the student model close to the teacher model.

Several studies find the problem that a large gap between the teacher and student models can impede effective knowledge distillation, and propose some solutions. \cite{mirzadeh2020improved} proposes the integration of a medium-sized teaching assistant model positioned between the teacher and student models. This framework enables the teaching assistant model to learn from the teacher model while allowing the student model to learn from the teaching assistant model. Consequently, this approach reduces the size disparity between the two models and enhances the distillation effect. \cite{park2021learning} suggests training a teacher model that is more compatible with student models. They achieve this by incorporating some of the structure of the student model into the teacher model. By doing so, the teacher model gains an understanding of the student model's structure, thus generating features that facilitate the learning process of the student model. However, it is important to note that these methods are applicable only when dealing with teacher and student models of different sizes within the same modality. Therefore, addressing significant gaps between modalities remains a challenging task.

In the field of cross-modal knowledge distillation research, there is currently a focus on transferring knowledge between similar modalities. \cite{gupta2016cross} proposes a method that utilizes a model of RGB images as a teacher model to guide the training of depth and optical flow image models. Similarly, \cite{tian2019contrastive} introduces contrastive representation distillation, which applies cross-modal distillation from RGB images to depth images. Some techniques that distill knowledge between disparate modalities often rely on the teacher modality to provide labels for the student model in an unsupervised setting. \cite{albanie2018emotion} designs a method to provide label information for unlabelled speech emotion recognition models using pre-trained facial emotion recognition models as teacher models. \cite{inoue2021teacher} leverages a face recognition model to generate positive and negative pairs for unlabelled speech data, and uses metric learning to train the speaker recognition model. In contrast to these unsupervised approaches, our work explores supervised cross-modal distillation. Building upon this, \cite{zhang2021knowledge} employs a multi-modal teacher consisting of face and speech information to guide the training of a single-modal student. However, they encounter challenges due to the significant gap between the speech and teacher systems, which hinder performance improvement. \cite{chen2021distilling} propose a cross-modal distillation method based on contrastive learning, but we found that it is not applicable in all scenarios. The goal of this paper is to find methods that enable more generalized knowledge transfer between widely differing modalities.

\subsection{Speaker Recognition}

Speaker recognition, an important branch of biometrics, identifies people by their voice. It can be categorized into speaker identification and verification, the former is carried out in a closed set to determine the identity of the person to whom the voice belongs, and the latter is carried out in an open set to determine whether the two voices belong to the same person, which is more commonly used. The traditional approach to speaker recognition involves the utilization of I-vector \cite{dehak2010front}.

The field of speaker recognition has made remarkable progress in recent years, particularly with the advent of deep neural networks. \cite{snyder2018x} introduces an early deep learning model called X-Vector, which exhibits superior performance compared to traditional methods. \cite{nagrani2017voxceleb} proposes the VGGM model, which further improves the results in speaker recognition. To improve the pooling process, \cite{okabe2018attentive} develops the attentive statistics pooling method, incorporating attention weighting and additional statistics. \cite{chung2020defence} conducts a comprehensive analysis of various loss functions and identifies the angular prototypical loss as the most effective for speaker recognition tasks. \cite {desplanques2020ecapa} proposes the ECAPA-TDNN, a highly robust and powerful model for speaker recognition.

In terms of the dataset, Chung et al. propose the VoxCeleb1 \cite{nagrani2017voxceleb} and VoxCeleb2 \cite{chung2018voxceleb2} datasets, which is one of the most commonly used speaker recognition datasets, and establish a new benchmark using the ResNet model \cite{he2016deep}. Additionally, \cite{9054017} introduces the CN-Celeb dataset, a comprehensive collection of Chinese data that covers various scenarios.

Some researchers endeavor to integrate face images and audio in the context of audio-visual speaker recognition. Some studies have explored the correlation between face images and speech, \cite{nagrani2018seeing} proposing a classification network that demonstrates that neural networks can achieve a match between the two modalities that is comparable to human performance, which also demonstrates the existence of rich modality-sharing features between the two modalities. Other researchers mix the two modalities to achieve better recognition performance. \cite{sari2021multi} advances the idea of acquiring joint audio-visual embeddings and employing cross-modal verification. Furthermore, \cite{qian2021audio} propose and evaluate various multi-modal fusion techniques. Nevertheless, it is important to acknowledge that the practical use of these methods may face a limitation due to the unavailability of face images. In contrast, our approach solely relies on speech input during testing.

\subsection{Image Classification}

Image Classification is a fundamental task in computer vision, aimed at identifying the category to which an image belongs. The Cifar series dataset, developed by \cite{krizhevsky2009learning}, represents an early example of an image classification dataset. Despite its smaller size, consisting of 60,000 images, it has played a significant role in the field. In contrast, ImageNet \cite{ILSVRC15} has emerged as the most widely used large-scale image classification dataset, boasting millions of images. Notably, ImageNet organizes annual competitions that center around image classification, attracting considerable attention from researchers. Over the course of these competitions, numerous effective neural networks for image classification have been proposed. For instance, AlexNet \cite{krizhevsky2017imagenet} and VggNet \cite{simonyan2014very}, two earlier neural networks, feature fewer layers but have proven to be valuable in the field. ResNet \cite{he2016deep}, on the other hand, has become the go-to convolutional neural network due to its successful introduction of residual connections, which tackle the problem of gradient disappearance in deep networks and allow for significantly increased network depth. Building upon ResNet, SENet \cite{hu2018squeeze} incorporates a squeeze-excitation module to further enhance performance. 

In recent times, several researchers find that using textual modalities to supervise image models allows for the training of high-performance and strongly generalizable image classification models. Notably, a significant advancement in this domain is observed with the advent of CLIP \cite{radford2021learning}, a methodology that leverages extensive image-text paired data to supervise image training through textual modalities. More approaches are proposed later. CyCLIP \cite{NEURIPS2022_2cd36d32} aims to formalize consistency and the learning of geometrically consistent representations in both image and text space. xCLIP \cite{Zhou_2023_CVPR} introduce non-contrastive image-text pre-training and demonstrated that it facilitates zero-shot representation learning, and the combination with contrastive pre-train achieves the best results. DPL~\cite{DPL, song2025dpl++} reveals the advances brought about by the simultaneous optimization of images and labels. These methods utilize textual modal supervision to pre-train image models using large-scale data, and then fine-tune image models for downstream tasks such as classification. Instead, we explore the feasibility of introducing textual modality transfer knowledge in image classification model training directly without pre-train.

\begin{figure*}[ht]
  \centering
  \includegraphics[width=\linewidth]{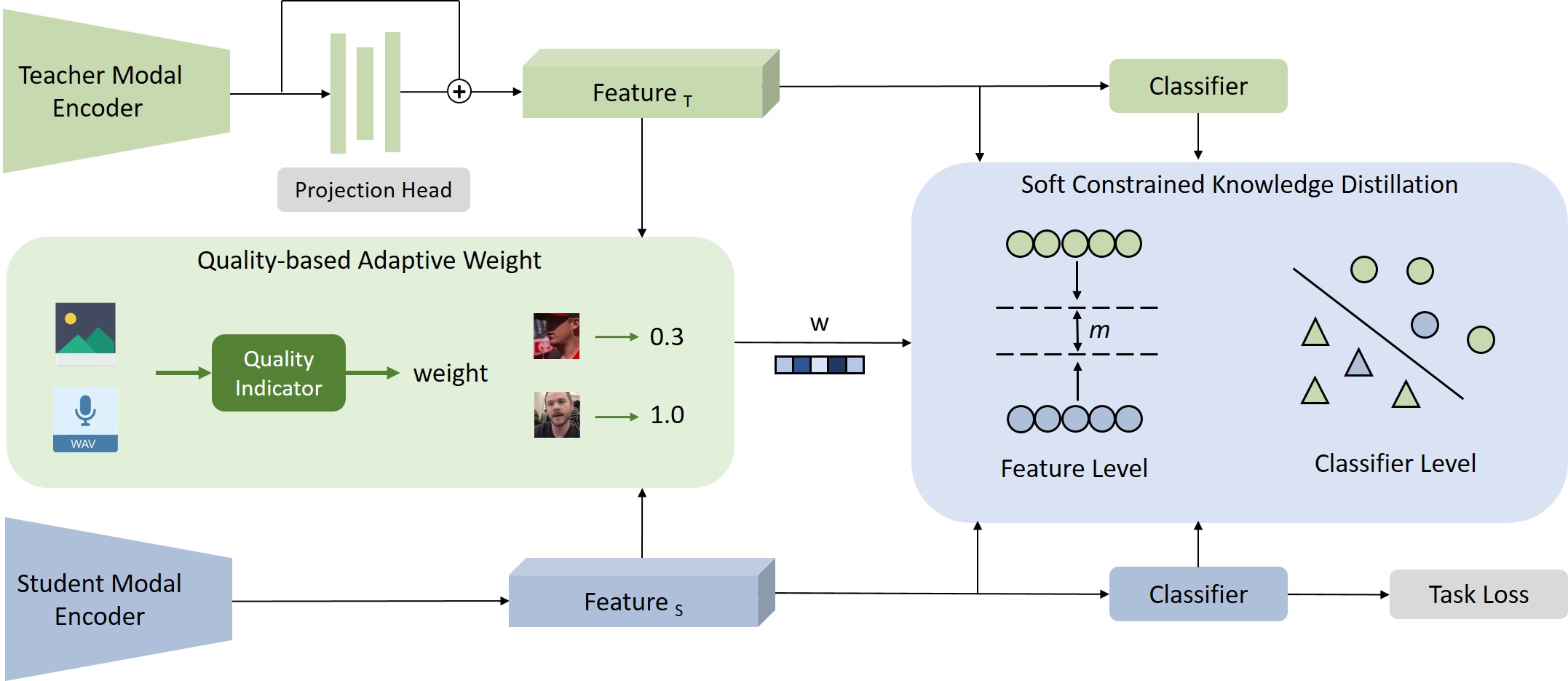}
  \caption{The overview of our cross-modal distillation framework where the teacher modality transfers discriminative knowledge to the student modality. The projection head extracts modality-share features from the teacher model to reduce the modal gap. Soft constrained knowledge distillation sets distillation constraints from the feature level and classifier level, which can avoid overfitting while transferring knowledge. The quality module can weigh the training samples based on the quantified data quality, leading to robust training.}
  \label{fig:overview}
\end{figure*}

\section{Method}
\subsection{Overview}

The traditional knowledge distillation methods often struggle when transferring knowledge between significantly different modalities, such as images and audio, primarily due to the large modality gap. This gap arises from the inherent differences in data representation and processing in various modalities, leading to inefficient knowledge transfer and potential overfitting. To address these challenges, we propose a comprehensive cross-modal knowledge distillation framework. Our framework is designed to bridge the modality gap by focusing on three key components: extraction of modality-shared features, implementation of soft constraints, and assessment of data quality. These components work synergistically to facilitate effective knowledge transfer between disparate modalities.

\textcolor{black}{The foundation of our approach lies in the extraction of features that are shared across modalities. To achieve this, we introduce a trainable projection head atop the teacher model. This projection head serves to transform the teacher's features into a space that is more aligned with the student's feature space. By doing so, we bridge the domain gap without the computational burden of fine-tuning the entire teacher model. This strategic alignment ensures that the student model can effectively learn from the teacher's knowledge, focusing on the shared aspects of the data. By adding a projection head to the teacher model, we create a bridge between the teacher's feature space and the student's, allowing for more effective knowledge transfer. This alignment is akin to translating the teacher's language into one the student can understand better.} Traditional hard constraints, such as L2 loss, are ill-suited for cross-modal distillation. They force the student model to mimic the teacher's features exactly, including modality-specific information that may be irrelevant or even detrimental to the student's task. In contrast, our framework employs soft constraints at both the feature and classifier levels. These soft constraints encourage the student to learn modality-shared features while allowing flexibility to ignore irrelevant modality-specific features. This approach prevents overfitting and ensures that the student model generalizes well across different modalities.

\textcolor{black}{Using soft constraints allows the student to learn general patterns without mimicking the teacher's idiosyncrasies, thus preventing overfitting to irrelevant features. The quality of input data plays a crucial role in the effectiveness of knowledge distillation. Low-quality data can mislead the learning process, leading to suboptimal performance. To mitigate this, our framework adaptively adjusts training objectives based on data quality. By prioritizing high-quality data and downweighting noisy or low-quality samples, the framework becomes more robust and reliable, ensuring that the student model learns from the most informative examples. Focusing on shared features enables the student model to generalize better across modalities, as it learns features relevant in both domains. The proposed framework is grounded in information theory, particularly the concept of mutual information. By aligning feature spaces through the projection head and employing soft constraints, we maximize the mutual information between the teacher and student models, focusing on shared knowledge. This alignment is akin to translating between different languages, where the projection head acts as a dictionary, facilitating effective communication.}

Formally, the inputs from the student and teacher modalities are first encoded by the encoder of the respective modality to obtain the features $F_S$ and $F_T$. The student feature $F_S$ learns knowledge from the teacher modality through the proposed soft constrained distillation loss, and this distillation loss has an adaptive weight $\mathbf{w}$ obtained from the quality of the input data to achieve robustness to data quality. The total loss is the combination of the task loss with a task head $\sigma$ and supervised by label information $y$ and the distillation loss, 

\begin{equation}
  \mathcal{L} = \mathcal{L}_{task}(\sigma(F_S), y) + \mathbf{w} \cdot \mathcal{L}_{distillation}(F_S, F_T)
\end{equation}

The following sections detail the projection head (Sec. \ref{sec:proj}), soft constrained knowledge distillation (Sec. \ref{sec:sckd}), and quality-based adaptive weight (Sec. \ref{sec:qaw}).

\subsection{Projection Head}
\label{sec:proj}
In cross-modal scenarios, applying a fixed teacher model as in general knowledge distillation methods usually does not yield good results. 
It results from the large domain gap between the two modalities in cross-modal scenarios. Appropriate adjustments to the teacher model to extract modality-shared features are necessary and can also make the student model easier to fit. Fully fine-tuning the entire teacher model incurs a large arithmetic overhead and can easily reduce the performance of the teacher model, so we add a trainable MLP projection head with a small number of parameters to the fixed teacher model for parameter-efficient fine-tuning. 

Assume that the input $I_T$ of the teacher modality is encoded by the fixed teacher model encoder $Encoder_T$ to obtain embedding $E_T$, which is fine-tuned by a trainable MLP to get the $E_{T}^{'}$.

\begin{equation}
  E_T = Encoder_T(I_T)
\end{equation}

\begin{equation}
  E_{T}^{'} = MLP(E_T)
\end{equation}


To make use of the features $E_{T}$, then we add a skip connection to fuse the feature $E_{T}$ and $E_{T}^{'}$ via a hyperparameter $\alpha$.  The final feature of teacher model $F_T$ can be obtained by

\begin{equation}
  F_T = \alpha \cdot E_T + (1-\alpha) \cdot E_{T}^{'}
  \label{equ:feature_T}
\end{equation}

\subsection{Soft Constrained Knowledge Distillation}
\label{sec:sckd}
Distillation loss  guides the student  to learn discriminative features from the teacher. 
The traditional distillation loss, e.g., $l$2 loss, conducts \emph{hard} constraint to force the student to be exactly the same as the teacher, easily leading to overfitting in cross-modal scenarios due to the big domain gap. 
To address this, in this work, 
we design \emph{soft} constrained knowledge distillation loss at the feature level and classifier level respectively.

\subsubsection{Feature Level}
\begin{figure}[ht]
  \centering
  \includegraphics[width=0.7\linewidth]{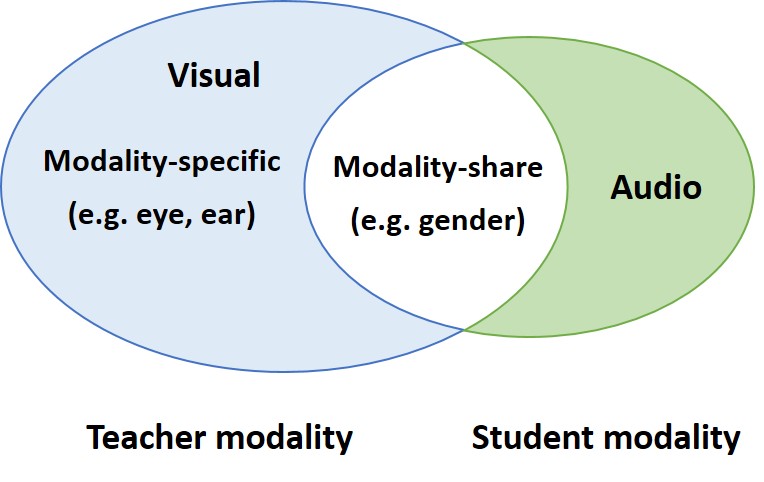}
  \caption{Knowledge transfer can lead to overfitting if modality-specific information is transferred.}
  \label{fig:venn}
\end{figure}

Knowledge transfer through intermediate features or the output of the last layer of the model is a common practice in knowledge distillation. The general distillation loss usually aims to set some kind of feature constraint to maximize the mutual information between student features and teacher features. However, in cross-modal scenarios, several new challenges arise. There are modality-shared features and modality-specific features between two different modalities, as illustrated in Fig. \ref{fig:venn}. For example, in the identity recognition task, features such as gender and age are shared between the face image and speech modalities. However, features such as eye and ear are unique to the visual modality and cannot be directly acquired from the audio modality. Consequently, the upper bound of mutual information is determined by the modality-shared features between the two modalities, and continued expansion will lead to the problem of overfitting irrelevant features and failing to achieve optimal results.

\textcolor{black}{In order to reduce the domain gap between the teacher and student models, we introduce a trainable projection head on the teacher model. This projection head allows the teacher model to project its features into a space that is more aligned with the student model's feature space, thereby facilitating the transfer of modality-shared knowledge. By fine-tuning only a small number of parameters in the projection head, we can efficiently adjust the teacher model's output to better match the student model without incurring significant computational overhead. This parameter-efficient fine-tuning helps in reducing the domain gap and makes the student model easier to train. Hard alignment, such as using $l$2 loss, forces the student model to mimic the teacher model's features exactly. However, in cross-modal scenarios, this can lead to the student model learning modality-specific features that are irrelevant to its own modality. By using soft constraints, we allow the student model to learn only the modality-shared features, as these are the features that are relevant across different modalities. Soft constraints provide a more flexible alignment that focuses on the shared aspects of the features, thereby preventing the student model from overfitting to modality-specific information. This approach ensures that the student model learns the most relevant and transferable knowledge from the teacher model, leading to better generalization and performance.}

This makes it necessary to make adjustments to the existing distillation constraints. Existing feature distillation methods generally maximize the mutual information between teacher and student models through contrastive learning \cite{chen2021distilling} or direct optimization of the distance between the teacher and student \cite{2015FitNets, chen2022knowledge}. Contrastive learning approaches typically use the same categories of teacher features and student features as positive pairs and different categories as negative pairs. However, in cross-modal scenarios, especially between modalities with weak associations, positive and negative pairs may contain close modality-share features, and empirically this approach does not work well for knowledge transfer. Therefore, we choose the non-contrastive approach and optimize the distance between paired teacher and student features only, as shown in Eq. (\ref{equ:feature}), $\mathcal{L}_{dis}$ is the  metric like $l2$ distance, cosine distance, etc.

\begin{equation}
  \mathcal{L}_{feature} = \mathcal{L}_{dis}(F_S, F_T) \label{equ:feature}
\end{equation}

However, setting this constraint directly still does not solve the overfitting problem, as the teacher features and student features cannot be fully aligned due to the presence of modality-specific features. Inspired by this,  we have devised a simple way, motivated by metric learning \cite {schroff2015facenet} which usually uses a margin to separate positive and negative pairs, we introduce a margin $m$ to  loosely align the teacher and student. In this way, we do not push the student forward to be exactly the same as the teacher. Instead, we use a margin $m$ to bound the maximal similarity between teacher and student. This margin brings a mechanism that can potentially ask the student to learn the discriminative information from the teacher and effectively avoid overfitting.

In this way, our method can easily adapt to mainstream distillation methods. We introduce a hyperparameter $m$ to construct the distillation loss, rewriting the Eq. (\ref{equ:feature}) as

\begin{equation}
  \mathcal{L}_{feature} = \left\lfloor \mathcal{L}_{div}(F_S, F_T) - m \right\rfloor 
\end{equation}

where $\left\lfloor\right\rfloor$ means cut down to 0, equivalent to $max(\cdot, 0)$. Clearly, 
when the distance between the two modalities is reduced to $m$, the loss becomes 0. 

\subsubsection{Classifier Level}
In a general classification task, features encoded by the encoder are passed through a classifier, which then calculates the cross-entropy loss based on the labels. As shown in Eq. (\ref{equ:loss_ce}), where $L$ is the logits obtained by the classifier, $y$ is the label information.

\begin{equation}
\mathcal{L}_{CE}(L, y)=-log(\frac{exp(L_y)}{\sum_{y'} exp(L_{y'})})
\label{equ:loss_ce}
\end{equation}

Classifier-based features, i.e. the logits, are also commonly used in knowledge distillation and typically encode a higher level of semantics compared to intermediate features. A common practice of knowledge distillation requires the logits distribution of the student model to be similar to that of the teacher model usually via a KL divergency loss. However, in cross-modal scenarios, the distribution of logits cannot be exactly the same in different modalities due to big modality gap, and empirically this does not lead to promising performance. 

\begin{figure}[ht]
  \centering
  \includegraphics[width=\linewidth]{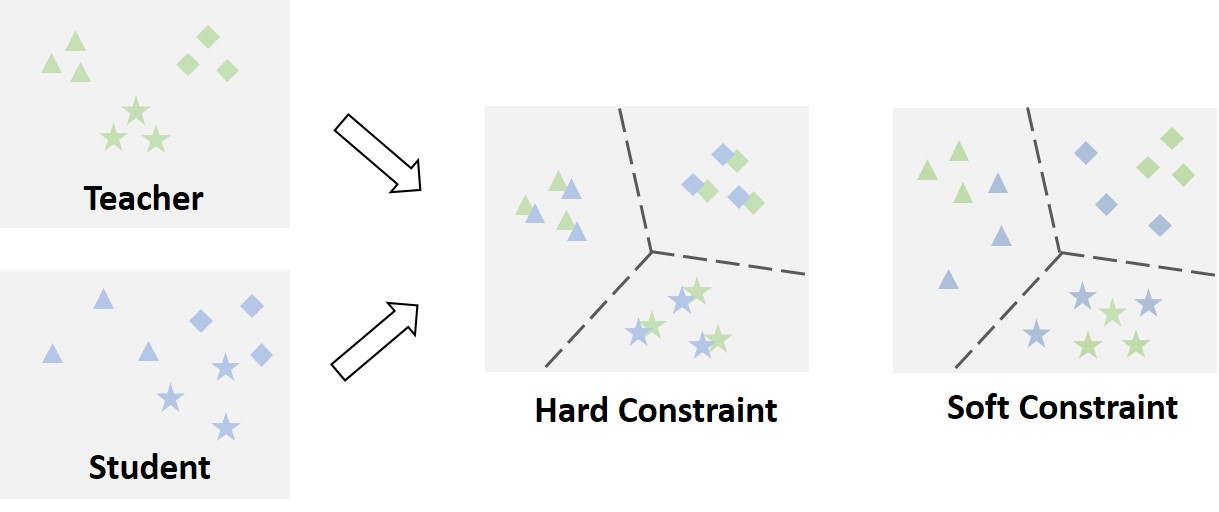}
  \caption{Illustration of the classifier level distillation. Different shapes indicate different categories and dash lines indicate decision boundary.}
  \label{fig:cla}
\end{figure}

Recent studies \cite{tung2019similarity, park2019relational} find cross-sample relationship captures the distribution of samples in classifier feature space, aligning this distribution between the teacher and student showing desirable performance in single-modality distillation. However, directly forcing the student to have the same distribution as the teacher will lead to overfitting in our cross modality scenario. 
To reduce this overfitting and capture the cross-sample distribution, we are inspired to  ask the teacher and student to share the classifier in a \emph{soft} way. As shown in Fig. \ref{fig:cla}, we force the samples of one  class of both the teacher and  student to stay within the classifier's decision boundary of that class, allowing the student model to well learn the  sample distribution in latent space under a soft constraint. 
This \emph{soft}  classifier sharing strategy clearly differs from  the hard constraint which forces the samples of the teacher and student to be exactly the same.

We empirically find that transferring knowledge across modalities, especially between widely differing modalities, poses a greater challenge compared to general knowledge distillation. This emphasis arises from several reasons:

\textcolor{black}{
(1) \textbf{Discrepancies in Input Modalities and Model Structures:} Cross-modal knowledge distillation encounters significant gaps due to the inherent differences in how different modalities represent and process data. For instance, images are typically represented as dense, high-dimensional arrays, while speech signals are represented as time-series data with varying frequency components. These differences lead to substantial disparities in model architectures and feature representations, making it difficult to directly transfer knowledge between modalities. Studies \cite{mirzadeh2020improved, park2021learning} have indicated that knowledge distillation is less effective when the disparity between the teacher and student models is substantial, highlighting the need for tailored approaches to address these challenges.
}

\textcolor{black}{
(2) \textbf{Modality-Specific and Shared Features:} Different modalities contain both \textbf{modality-shared features} and \textbf{modality-specific features}, which complicates the alignment process. For example, in the speaker recognition task, features such as gender and age are shared between face images and speech, while features such as pitch and timbre are specific to speech. During the distillation learning process, there is a risk of overfitting to irrelevant modality-specific features, which can degrade the performance of the student model. This issue underscores the need for alignment strategies that focus on modality-shared features while minimizing the influence of modality-specific features.
}

\textcolor{black}{
(3) \textbf{Varying Data Quality Across Modalities:} The quality of input data significantly impacts model performance, and this is particularly problematic in cross-modal distillation. The teacher and student models often utilize different input data, leading to misaligned input quality that may interfere with the distillation training process. For instance, in a scenario where the teacher model is trained on high-quality images, while the student model is trained on noisy or low-resolution audio, the distillation process can be compromised. This highlights the need for mechanisms to adaptively account for differences in data quality across modalities, ensuring that the distillation process remains robust and effective.
}


Specifically, we first mix the features of the data and labels from the two modalities encoded by their respective encoders. Assuming the batch size is $b$, $2b$ samples and the corresponding labels can be obtained.

\begin{equation}
\mathbf{F}=\{F_S, F_T\},\quad \mathbf{y}=\{y_S, y_T\}
\end{equation}

We then pass the combined features through a shared classifier $w$, which maps the two modalities into the same space without pursuing \emph{hard} agreement and is an effective way of soft constraint.

\begin{equation}
    \mathbf{L} = w \cdot \mathbf{F}
\end{equation}

In our practice, we find if the correlation between modalities is strong, the logits of the two modalities can be further aligned by adding a hard constraint like $l2$ to the logits of the two modalities. 
Thus, our final classifier-level loss is the combination of the hard and soft losses weighted by 
a hyperparameter $\beta$. 
The total distillation loss at the classifier level is

\begin{equation}
\mathcal{L}_{classifier}=\mathcal{L}_{CE}(\mathbf{L}, \mathbf{y}) + \beta \cdot \mathcal{L}_{dis}(L_s, L_t)
\label{equ:loss_cla}
\end{equation}

\subsection{Quality-based Adaptive Weight}
\label{sec:qaw}
The performance of a model can be greatly influenced by the quality of the input data. In cross-modal distillation scenarios, where the teacher model and the student model have different inputs, the quality of the input data for the two modalities can affect the performance of both. This misalignment in data quality can further disrupt the distillation training process. It is not hard to understand this. For example, if the input quality of the teacher modality is poor (e.g. blurry faces), the teacher potentially misleads the student. It motivates us to weigh the samples by data quality.

For a modality, the quality of the input data directly impacts the richness of the features that can be extracted, low quality data may interfere with distillation training. If the quality of the input data is poor and it contains few features, distillation constraints may lead to the problem of overfitting teacher features or being misled by teacher features. This means that the distillation loss weight for low quality data should be reduced.

For quantifying sample quality, there exist many specialized models that can quantify data quality, however, these models are usually computationally expensive. We use a simple way to quantify input quality by using $l2$ feature norms to obtain the quantized quality $Q_i$, which are proven to be highly correlated with input quality \cite{kim2022adaface}.
    
We then map the quantized sample quality to the range of variation in weights, assuming that the data quality of a sample is $Q_i$, its mean is $\mu_q$, its variance is $\sigma_q$, and the base weights are $w_{base}$. The redistributed weights are as shown in Eq. (\ref{equ:weight}), with low-quality samples assigned lower weights, $\mu_q$ and $\sigma_q$ can be calculated by moving average to increase stability, and $h$ is a hyperparameter to control the range of variation of weights. Typically, $w_{base}$ is 1 and $h$ is $1/3$.

\begin{equation}
  w_i = w_{base} + \frac{Q_i - \mu_q}{\sigma_q / h}
  \label{equ:weight}
\end{equation}

\section{Experiments}
We conduct experiments on cross-modal knowledge transfer, specifically focusing on image, speech, and text as commonly used but widely differing modalities. The downstream tasks for these experiments involve speaker recognition, with the teacher modality being the facial image and the student modality being speech, as well as image classification, with the teacher modality being text consisting of labels and the student modality being images. The implementation details and results of each experiment are specified below.

\begin{table*}[ht]
    \centering
    \caption{\textcolor{black}{Knowledge distillation comparisons using ResNet34. O and H represent the original and hard test set of VoxCeleb1, respectively. The right 3 columns are the test results after adding noises with different levels.}}
    \resizebox{\linewidth}{!}{
    \begin{tabular}{ccccccc}
    \hline
    method & EER (O) & minDCF (O) & EER (H) & EER ($\Delta_{db}=15$) & EER ($\Delta_{db}=10$) & EER ($\Delta_{db}=5$) \\
    \hline
    \hline
    ResNet34 (w/o dist.) & 1.71\% & 0.205 & 3.32\% & 2.78\% & 3.71\% & 5.61\% \\
    \hline
    \hline
    KD \cite{hinton2015distilling} & 1.70\% & 0.208 & 3.60\% & 2.74\% & 3.73\% & 5.52\% \\
    FitNet \cite{2015FitNets} & 1.81\% & 0.233 & 3.68\% & 2.82\% & 3.62\% & 5.37\% \\
    PKD \cite{passalis2018probabilistic} & 1.67\% & 0.204 & 3.21\% & 2.71\% & 3.57\% & 5.36\% \\
    SP \cite{tung2019similarity} & 2.18\% & 0.259 & 4.11\% & 3.50\% & 4.62\% & 6.97\% \\
    ICKD \cite{liu2021exploring} & 1.75\% & 0.209 & 3.35\% & 2.88\% & 3.70\% & 5.48\% \\
    CCL \cite{chen2021distilling} & 2.58\% & 0.368 & 5.79\% & 3.63\% & 4.36\% & 5.84\% \\
    \hline
    \hline
    \textcolor{black}{Ours (feature)} & \textbf{1.51\%} & 0.178 & 3.21\% & \textbf{2.37\%} & 3.26\% & 4.97\% \\
    \textcolor{black}{Ours (classifier)} & 1.63\% & 0.186 & 3.29\% & 2.65\% & 3.52\% & 5.25\% \\
    \textcolor{black}{Ours (feature+quality)} & 1.54\% & \textbf{0.165} & 3.13\% & \textbf{2.37\%} & \textbf{3.20\%} & \textbf{4.77\%} \\
    \hline
    \end{tabular}
    }
    \label{tab:dist}
\end{table*}

\subsection{Speaker Recognition}

\subsubsection{Implementation Details}
\textbf{Data Preprocessing.} During training, we use randomly cropped 2 seconds speech segments, without any other data enhancements. 40-dimensional filter-banks (Fbank) with a window of width 25ms and step 10ms are used as the input. For the visual modality, we take 1 frame from each video, crop out the face part, then align it, and finally scale it to $112 \times 112$ size as the input of the network. In the testing phase, only data from speech modality is used as inputs.

\textbf{Datasets.} This study employs the VoxCeleb2 dataset \cite{chung2018voxceleb2} with multimodal data of speech and images for training and evaluates its performance on the VoxCeleb1 dataset \cite{nagrani2017voxceleb}. Both datasets are sourced from YouTube. The VoxCeleb1 dataset consists of 1,251 speakers and more than 100,000 utterances, only has data for audio modality. Its original test set comprises 37,720 randomly selected speaker pairs, while the hard-set test set includes 552,536 pairs with consistent race and gender. In contrast, the VoxCeleb2 dataset encompasses over a million utterances from 6,000 different speakers and provides both audio and visual modalities. Notably, there are no overlaps in terms of speaker identities between the two datasets.

To test whether modality-share features are learned during distillation training, we also test the accuracy of cross-modal matching. 10,000 randomly selected triples in the test set of VoxCeleb2 containing either one audio and two face images (A-F) or one face image and two audios (F-A) are selected, and the model needed to choose the matched one between the two. An additional hard test set is set up where the gender is kept the same for both options.

In order to test the generalizability across datasets, experiments were conducted using the CN-Celeb dataset \cite{9054017}. This dataset encompasses various genres, such as interviews, singing, and movies, in the Chinese language. Notably distinct from the VoxCeleb training dataset, CN-Celeb provides an effective means to evaluate generalization. With its test set comprising 18,849 utterances from 200 speakers and offering 3,484,292 test pairs, the likelihood of chance occurrences is significantly reduced.

\textbf{Model.} For the speaker recognition model, We tested the advanced performance representative ResNet34 (with ASP \cite{okabe2018attentive} for aggregate temporal frames), which is the most commonly used and advanced models in speaker recognition. The loss function is a combination of angular prototypical loss \cite{chung2020defence} and cross-entropy loss. As for teacher model, we use a pre-trained face recognition model IR-50 with high performance taken from \cite{wang2021face}. \textcolor{black}{In addition, to test the effect of different performance teacher models on the results, we also trained a poorly performing face recognition teacher model on a subset of VoxCeleb2 using ResNet34.}

\textbf{Hyperparameters}. Unless otherwise stated, the mixing ratio of the projected heads in the experiments of this section is $\alpha=0.6$. For feature level distillation, we use the cosine distance as a metric, with a value of $cos(30^{\circ})$ for margin. For classifier level distillation, since the two modalities are weakly correlated, $\beta=0$. The effect of hyperparameters is explored in the ablation experiments below.

\textbf{Optimization.} For the optimization process, we adopt the Adam optimizer with an initial learning rate of 1e-3, which is reduced by 25\% every 3 epochs. To control overfitting, a weight decay of 5e-5 is utilized. The training data is organized into batches, each containing 100 speakers, with each speaker contributing 2 audio utterances. These settings are consistent with \cite{chung2020defence}. The model is trained for 36 epochs using an Nvidia RTX 3090 GPU. The training duration for X-Vector is approximately 9 hours while training ResNet34 requires around 2 days.

\textbf{Evaluation.} Ten 4-second temporal crops are sampled from each test segment for evaluation, and we calculate the distance between all possible pairs (10 × 10 = 100), and use the mean distance as the score. This is the same as \cite{chung2018voxceleb2,chung2020defence}. We report two most commonly used evaluation metrics in speaker recognition: the Equal Error Rate (EER) and the minimum value of $C_{det}$, the lower the better in both cases. EER is the rate at which both acceptance and rejection errors are equal. And $C_{det}$ can be calculated by

\begin{equation}
  C_{det} = C_{miss} \times P_{miss} \times P_{tar} + C_{fa} \times P_{fa} \times (1 - P_{tar})
\end{equation}

where we assume a prior target probability $P_{tar}$ of 0.01 and equal weights of 1.0 between misses $C_{miss}$ and false alarms $C_{fa}$.

To evaluate the robustness under noise, we incorporate the musan dataset \cite{musan2015} to enhance our test set. In this process, we merge a segment of noise with each test audio sample. The intensity of noise is determined by the difference in decibels between the original speech and the distorted speech, denoted as $\Delta_{db}$. Consequently, a greater $\Delta_{db}$ value corresponds to a lower amplitude of noise.

\begin{equation}
  \Delta_{db} = db_{audio} - db_{noise}
\end{equation}

\subsubsection{Results}
\textcolor{black}{We first test our proposed method on the VoxCeleb2 dataset using the high-performance and widely used baseline ResNet34 and compared it with existing state-of-the-art knowledge distillation methods. Note that not all distillation methods can be used for our task. For example, since the spatial locations of visual and speech modalities are not correlated, methods using spatial locations \cite{zagoruyko2016paying} for distillation cannot be applied. The results are shown in Table \ref{tab:dist}. In the right columns of the table, we add noise to the test set to evaluate the model's robustness to noise, with the noise increasing in order from left to right. The results show that these existing methods are difficult to achieve effective performance improvement in this scenario even if some methods work in other cross-modal scenarios. The difficulty of this task lies in the fact that the two modalities, face image and speech, are very different and have limited correlation, and existing knowledge distillation methods usually pursue the strong constraint that the teacher and student models are identical, which can easily bring about fitting difficulties and overfitting of modality-specific features and lead to a poor result. In contrast, our proposed methods introduce some different soft constraints, reduce overfitting and achieve an effective performance improvement. In noisy scenarios, the performance improvement of our method becomes more obvious, especially when combined with quality assessment, showing good robustness.}

\begin{table*}[ht]
    \centering
    \caption{Experiments using different teacher models and different amounts of data.}
    \resizebox{\linewidth}{!}{
    \begin{tabular}{cccccc}
    \hline
    method & data volume & Teacher model & EER (Teacher) & EER (Student) & minDCF (Student) \\
    \hline
    ResNet34 (w/o dist.) & 25\% & - & - & 3.61\% & 0.409 \\
    \hline
    Ours (feature) & 25\% & IR-50 & 0.64\% & 3.22\% & 0.373 \\
    Ours (classifier) & 25\% & IR-50 & 0.64\% & 3.15\% & 0.376 \\
    \hline
    Ours (feature) & 25\% & ResNet34 & 11.91\% & 3.23\% & 0.366 \\
    Ours (classifier) & 25\% & ResNet34 & 11.91\% & 3.30\% & 0.369 \\
    \hline
    \hline
    ResNet34 (w/o dist.) & 100\% & - & - & 1.71\% & 0.205 \\
    \hline
    Ours (feature) & 100\% & IR-50 & 0.64\% & 1.51\% & 0.178 \\
    Ours (classifier) & 100\% & IR-50 & 0.64\% & 1.63\% & 0.186 \\
    \hline
    Ours (feature) & 100\% & ResNet34 & 11.91\% & 1.57\% & 0.183 \\
    Ours (classifier) & 100\% & ResNet34 & 11.91\% &  1.66\% & 0.195 \\
    \hline
    \end{tabular}
    }
    \label{tab:voxceleb_different_teacher_data}
\end{table*}

\begin{figure}[ht]
  \centering
  \includegraphics[width=\linewidth]{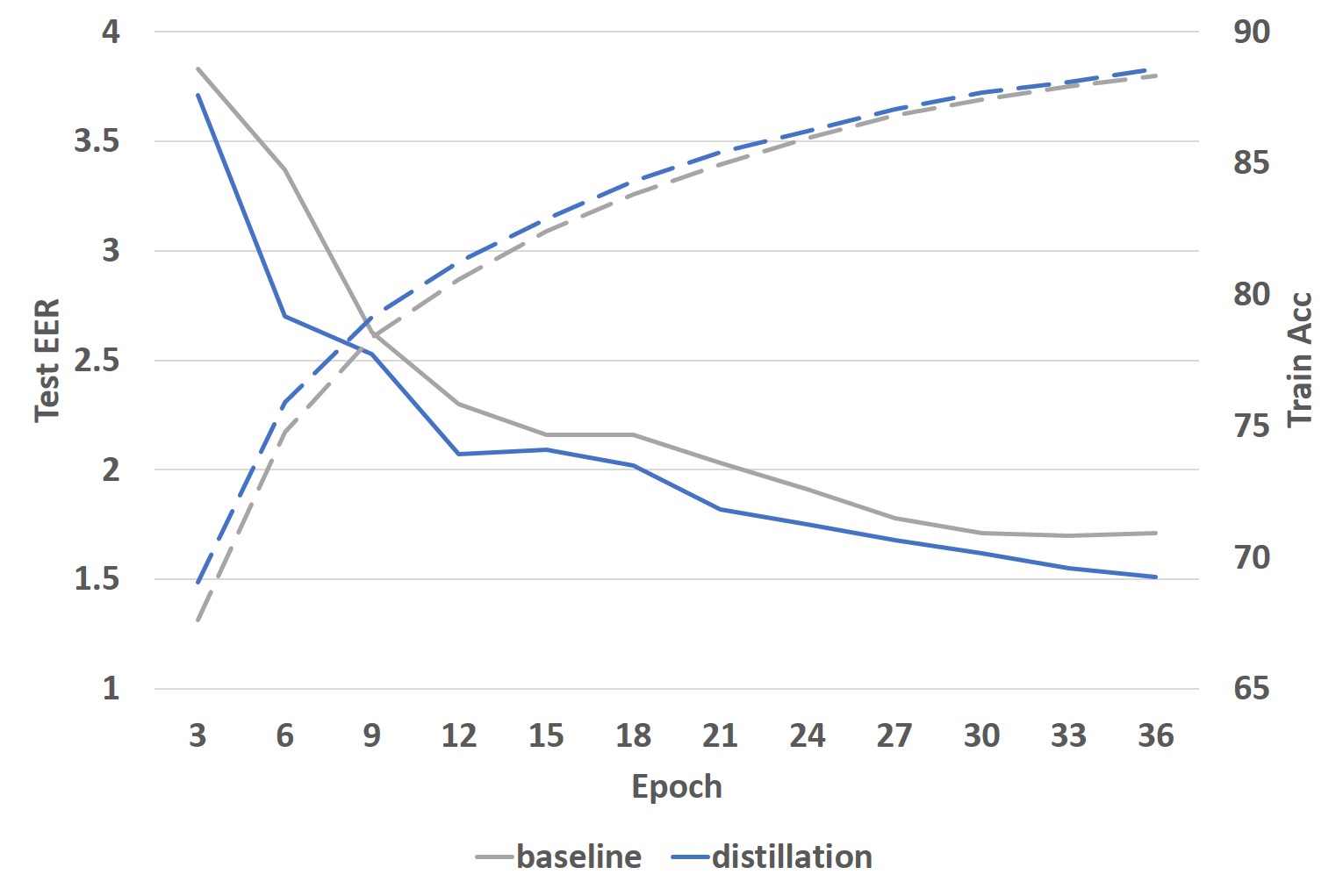}
  \caption{Training-set accuracy (dash line) and test-set EER (solid line) variation curves during model training.}
  \label{fig:vox_train}
\end{figure}

{The curve of accuracy in the training set and EER in the test set during the training process is shown in Fig. \ref{fig:vox_train}. It can be seen that there is a consistent improvement in accuracy and EER after applying our method compared to the baseline,  accelerating the convergence significantly.}

To evaluate the cross-dataset generalizability, we tested the ResNet34 model trained on the VoxCeleb2 dataset on the CN-Celeb dataset. These two datasets are very different in terms of video scenes and languages. Since the CN-Celeb dataset is collected from a variety of actual scenarios such as interviews, singing, vlog, etc., it is more difficult and has a higher error rate compared to VoxCeleb as shown in Table \ref{tab:generalizability}. Compared to the baseline, our proposed approach effectively improves the cross-dataset generalizability.

\begin{table}[ht]
    \centering
    \caption{Cross-dataset generalizability test. Trained on VoxCeleb2, tested on CN-Celeb.}
    \begin{tabular}{ccc}
    \hline
    model & EER & minDCF \\
    \hline
    ResNet34 (w/o dist.) & 14.49\% & 0.759 \\
    Ours & \textbf{13.03\%} & \textbf{0.713} \\
    \hline
    \end{tabular}
    \label{tab:generalizability}
\end{table}

\todo{We also test the performance in some practical scenarios. In real-world situations, the availability of a substantial amount of training data may be limited, so we assessed the performance of the method using a small amount of training data. We randomly extracted 25\% of the data from the VoxCeleb2 dataset. The results are presented in the upper part of Table \ref{tab:voxceleb_different_teacher_data}. With a small amount of data, our approach still achieves a robust  improvement, which is more significant compared to having sufficient data. Our findings reveal that when using the complete dataset (Table \ref{tab:dist}), feature-level distillation outperforms classifier-level distillation. However, when only 25\% of the dataset is used, classifier-level distillation performs equally well or better. This observation suggests that classifier-level distillation may be more suitable for small data, possibly due to its resemblance to a form of data augmentation.}

{We also explore the impact of different teacher models on results. As shown in Table \ref{tab:voxceleb_different_teacher_data}, we additionally use a face recognition teacher model with poorer performance. An interesting finding is that even when we replace a poorly performing teacher model, we are still able to improve the performance of the student model.  We believe that this
results from the fact that the 
data from different modalities can capture 
different perspectives and are highly complementary to each other.  Even if the teacher model performs poorly, the student model is still able to learn valuable knowledge.}



\begin{table}[ht]
    \centering
    \caption{Accuracy of cross-modal matching. The model needs to choose between two face images (A-F) or two speech (F-A) samples that match another modal identity. H indicates that the two options are of the same gender.}
    \resizebox{\linewidth}{!}{
    \begin{tabular}{ccccc}
    \hline
    method & A-F (O) & F-A (O) & A-F (H) & F-A (H) \\
    \hline
    chance & 50.00\% & 50.00\% & 50.00\% & 50.00\% \\
    FitNet \cite{2015FitNets} & 73.06\% & 73.72\% & 70.09\% & 71.40\% \\
    PKD \cite{passalis2018probabilistic} & 51.44\% & 52.12\% & 51.34\% & 51.79\% \\
    ICKD \cite{liu2021exploring} & 50.07\% & 50.77\% & 49.85\% & 49.84\% \\
    CCL \cite{chen2021distilling} & 74.87\% & 75.29\% & 70.38\% & 72.31\% \\
    \hline
    Ours (classifier) & 79.57\% & 80.86\% & 73.00\% & \textbf{75.95\%} \\
    Ours (feature) & \textbf{84.30\%} & \textbf{82.51\%} & \textbf{74.97\%} & 72.00\% \\
    \hline
    \end{tabular}
    }
    \label{tab:cross_modal_match}
\end{table}

\todo{To further validate whether modality-share features are learned during the distillation process, we conduct cross-modal matching tests using the distilled model. The model is required to select one of two faces to match the audio identity (A-F) or one of two audio segments to match the face image identity (F-A). For the original test set (O), the options were randomly selected, while for the hard test set (H), the options were of the same gender. The results of the test are shown in Table \ref{tab:cross_modal_match}. In the case of random selection, the accuracy is $50\%$. However, even though our model is not specifically trained for cross-modal matching, it achieves a high cross-modal matching accuracy, which may not be exceptionally high considering the limited correlation between the two modalities. This outcome proves that the student model indeed learned modality-share features during the training process. Notably, when the genders are the same, the matching accuracy remains significantly higher than random selection, suggesting that the model learned more potentially relevant features. Other existing knowledge distillation methods have poor cross-modal matching performance, which may be due to fitting difficulties associated with large cross-modal gaps that allow the model to learn few modality-share features. The CCL \cite{chen2021distilling} method utilizes contrastive learning and is better suited to learning modality-share features, but its low performance on the task itself in Table \ref{tab:dist} shows that it may have affected the learning of modality-specific features, whereas our method achieves both modality-share and modality-specific feature learning.}

\begin{figure}[ht]
  \centering
  \includegraphics[width=0.9\linewidth]{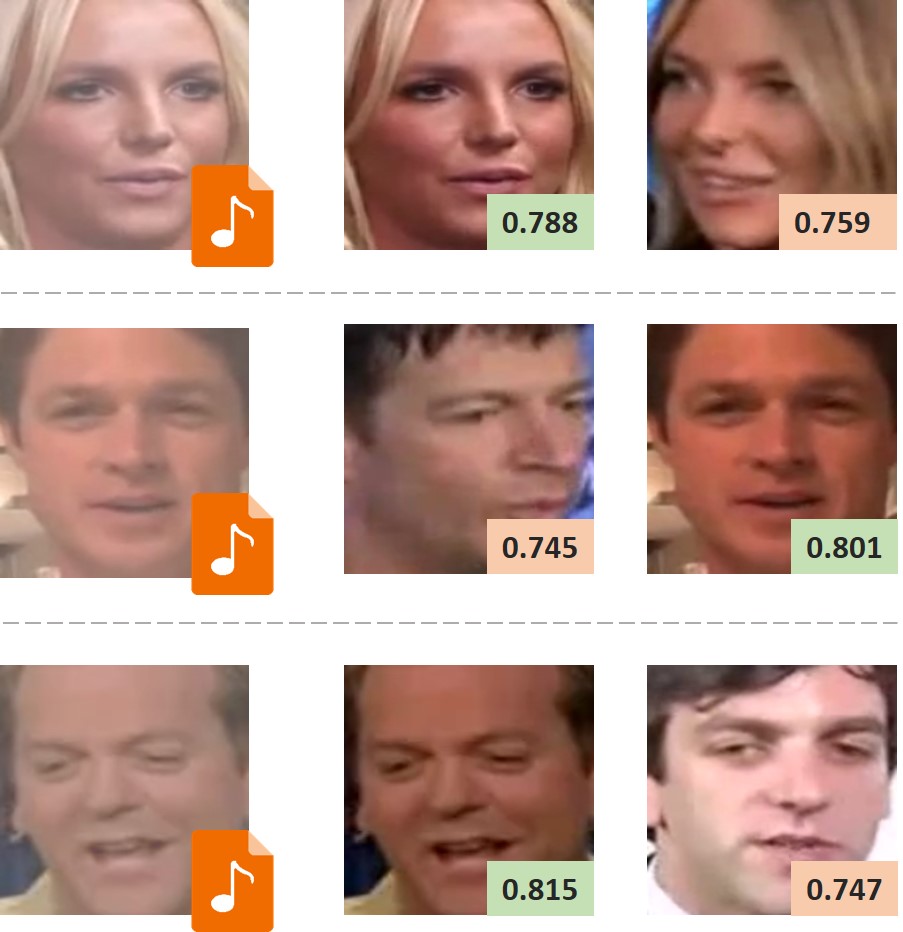}
  \caption{Some examples of cross-modal matching. Green indicates the correct answer and the number in the box is the cosine similarity between the modalities. Our model is able to match the correct answer among these similar faces.}
  \label{fig:cross_modal_match}
\end{figure}

\todo{Some cross-modal matching examples are shown in Fig. \ref{fig:cross_modal_match}. It can be seen that our model is still able to match correctly in these similar face images, indicating that rich modality-share knowledge has been learned.}

\begin{figure}[ht]
  \centering
  \includegraphics[width=0.8\linewidth]{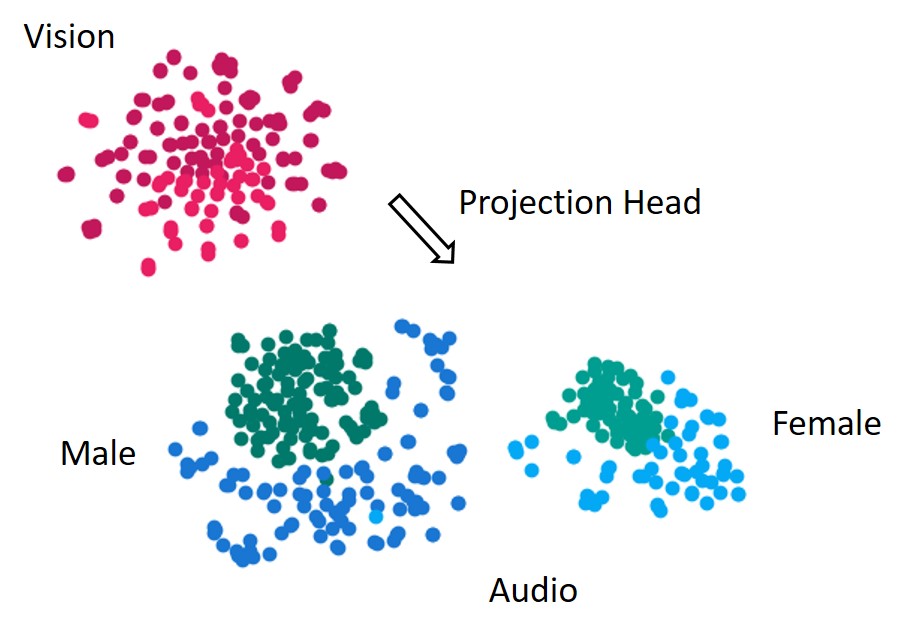}
  \caption{Features extracted by different models. 
  Color red, green, and blue represent the features from face model, projection head, and speech model respectively. 
  Dark and light colors represent male and female respectively.}
  \label{fig:visualization}
\end{figure}

To understand the features we learned, we visualize the features extracted by the face model, its projection head, and the speech model after applying our method. To achieve these features, we use 200 samples randomly selected from the test set. We conduct dimensionality reduction by t-SNE on these features. From Fig. \ref{fig:visualization}, we can see that there is a large domain gap between the two modalities of face (red) and speech (blue), which makes it difficult to align them directly. After the mapping of the projection head (green), the distance between the face and speech features is reduced, making the teacher-student distillation easier. In addition, the features extracted by the projection head and speech model do not completely overlap because our method uses a margin to avoid overfitting. Moreover, based on the annotation of gender (dark), it can be seen that the features mapped by the projection head well indicate gender, which is a shared attribute in face and speech, justifying our assumption that face and speech are highly correlated in feature space.

\subsubsection{Abation Study}
Then we conduct abation studies to verify the effectiveness of different components of our methods. We perform this study using ResNet34 on VoxCeleb1.

\textbf{Effect of projection head hyperparameter $\alpha$.} The face embedding from a pre-trained face model might have a big discrepancy against speech embedding. We propose a learnable projection head to reduce this discrepancy with the expectation that the feature from the projection head can more easily align with the speech feature. This feature itself can degrade the face recognition performance of the original feature. Thus, we introduce a hyperparameter $\alpha$ in Eq. (\ref{equ:feature_T}) to balance these two features. From Table \ref{tab:proj}, if $\alpha$ is too small, it means we mainly rely on the projected feature, and the face recognition performance might be degraded greatly; If $\alpha$ is too large, we mainly use the original face embedding,  which is very far from speech feature. It will cause difficulty for teacher-student distillation. Neither is the best choice, choosing an intermediate value of 0.6 gives the best results.

\begin{table}[ht]
    \centering
    \caption{Effect of mixing ratio of projection head.}
    \begin{tabular}{ccc}
    \hline
    $\alpha$ & EER & minDCF \\
    \hline
    0.2 & 1.68\% & 0.210 \\
    0.4 & 1.67\% & 0.195 \\
    0.5 & 1.52\% & 0.183 \\
    0.6 & \textbf{1.51\%} & \textbf{0.178} \\
    0.8 & 1.68\% & 0.199 \\
    \hline
    \end{tabular}
    \label{tab:proj}
\end{table}

\textbf{Effect of distillation hyperparameter $m$.} The margin $m$ controls the degree of similarity that the teacher and student should achieve. We use feature-based distillation for this experiment. From Table \ref{tab:margin}, if $m$ is 
too large, the constraints are too loose and the supervision is weak; If the $m$ is too small, it may lead to over-fitting. The best performance is achieved by $cos(30^{\circ})$. 

\begin{table}[ht]
    \centering
    \caption{Effect of margin values.}
    \begin{tabular}{ccc}
    \hline
    $m$ & EER & minDCF \\
    \hline
    0 & 1.66\% & 0.188 \\
    $cos(10^{\circ})$ & 1.62\% & 0.180 \\
    $cos(20^{\circ})$ & 1.71\% & 0.197 \\
    $cos(30^{\circ})$ & \textbf{1.51\%} & \textbf{0.178} \\
    $cos(40^{\circ})$ & 1.64\% & 0.197 \\
    $cos(50^{\circ})$ & 1.60\% & 0.189 \\
    \hline
    \end{tabular}
    \label{tab:margin}
\end{table}

\subsection{\todo{Image Classification}}

\subsubsection{Implementation Details}

\textbf{Data Preprocessing.} For the input image, during training, the image is randomly cropped and horizontally flipped to a size of $224 \times 224$ and normalized. During testing, the images are cropped centrally to $224 \times 224$ and then normalized. For the text modal, we randomly select one of the 80 templates (e.g. a photo of an object) given by CLIP \cite{radford2021learning} and format the category labels to form the input text. In the test phase, only the data of the image modality is used as input.

\textbf{Datasets.} We use two of the most commonly used datasets for image classification tasks for training and evaluation, Cifar100 \cite{krizhevsky2009learning} and Tiny-Imagenet \cite{le2015tiny}. The Cifar100 dataset contains 100 categories with 600 images per category, including 500 images in the training set and 100 images in the validation set. The Tiny-Imagenet is a larger dataset, a subset of ILSVRC2012, containing 200 categories, each category containing 500 training images and 50 validation images.

\textbf{Model.} For the image classification network, we use the most commonly used ResNet model without pre-train. For the text teacher network, we use the text encoder from the large-scale pre-trained CLIP ViT-B/16 \cite{radford2021learning} model.

\textbf{Hyperparameters}. Unless otherwise stated, the mixing ratio of the projected heads in the experiments of this section is $\alpha=0.6$. For feature-level distillation, we use the $l2$ distance as a metric, ResNet18 has a margin of $0.3^2$ and ResNet34 has a margin of $0.2^2$. For classifier level distillation, $\beta=1$. The effect of hyperparameters is explored in the ablation experiments below.

\textbf{Optimization.} We use the SGD optimizer with a momentum of 0.9 and a weight decay of 1e-4. Training is performed for a total of 90 epochs, with an initial learning rate of 0.1 and decreasing by 0.1 every 30 epochs. And the batch size is 32. We trained the model on a Nvidia RTX 2080Ti GPU, which takes about 4 hours for Cifar100 and 7 hours for Tiny-Imagenet.

\subsubsection{Results}

We initially conduct experiments on the Cifar100 dataset utilizing ResNet18 and ResNet34 models, as shown in Table \ref{tab:cifar}. This task is somewhat simpler compared to speaker recognition because the two modalities are more strongly correlated and the text consisting of labeled information contains the features needed for classification. However, the difficulty still exists, directly applying the loss of CLIP to the image classification task is not helpful for performance improvement, some methods can slightly improve the performance, but also very limited. In contrast, our approach effectively improves performance in this scenario, demonstrating its wide range of applicability. The ResNet18 model outperforms the baseline of the ResNet34 model after distillation by our method, which can effectively save arithmetic power. Classifier-level distillation performs better in this task, which again suggests that it is suitable for smaller datasets.

\begin{table}[ht]
    \centering
    \caption{Experimental results of ResNet18 and ResNet34 on Cifar100}
    \resizebox{\linewidth}{!}{
    \begin{tabular}{ccc}
    \hline
    method & Acc (ResNet18) & Acc (ResNet34) \\
    \hline
    Baseline (w/o dist.) & 75.43\% & 77.07\% \\
    \hline
    FitNet \cite{2015FitNets} & 75.97\% & 77.65\% \\
    SP \cite{tung2019similarity} & 73.42\% & 74.69\% \\
    ICKD \cite{liu2021exploring} & 75.29\% & 76.78\% \\
    PKD \cite{passalis2018probabilistic} & 76.10\% & 76.80\% \\
    CCL \cite{chen2021distilling} & 76.05\% & 77.36\% \\
    CLIP \cite{radford2021learning} & 75.64\% & 76.74\% \\
    \hline
    Ours (feature) & 76.28\% & 78.33\% \\
    Ours (feature+quality) & 76.57\% & 78.61\% \\
    Ours (classifier) & \textbf{77.44\%} & \textbf{79.14\%} \\
    \hline
    \end{tabular}
    }
    \label{tab:cifar}
\end{table}



\begin{figure}[ht]
  \centering
  \includegraphics[width=0.95\linewidth]{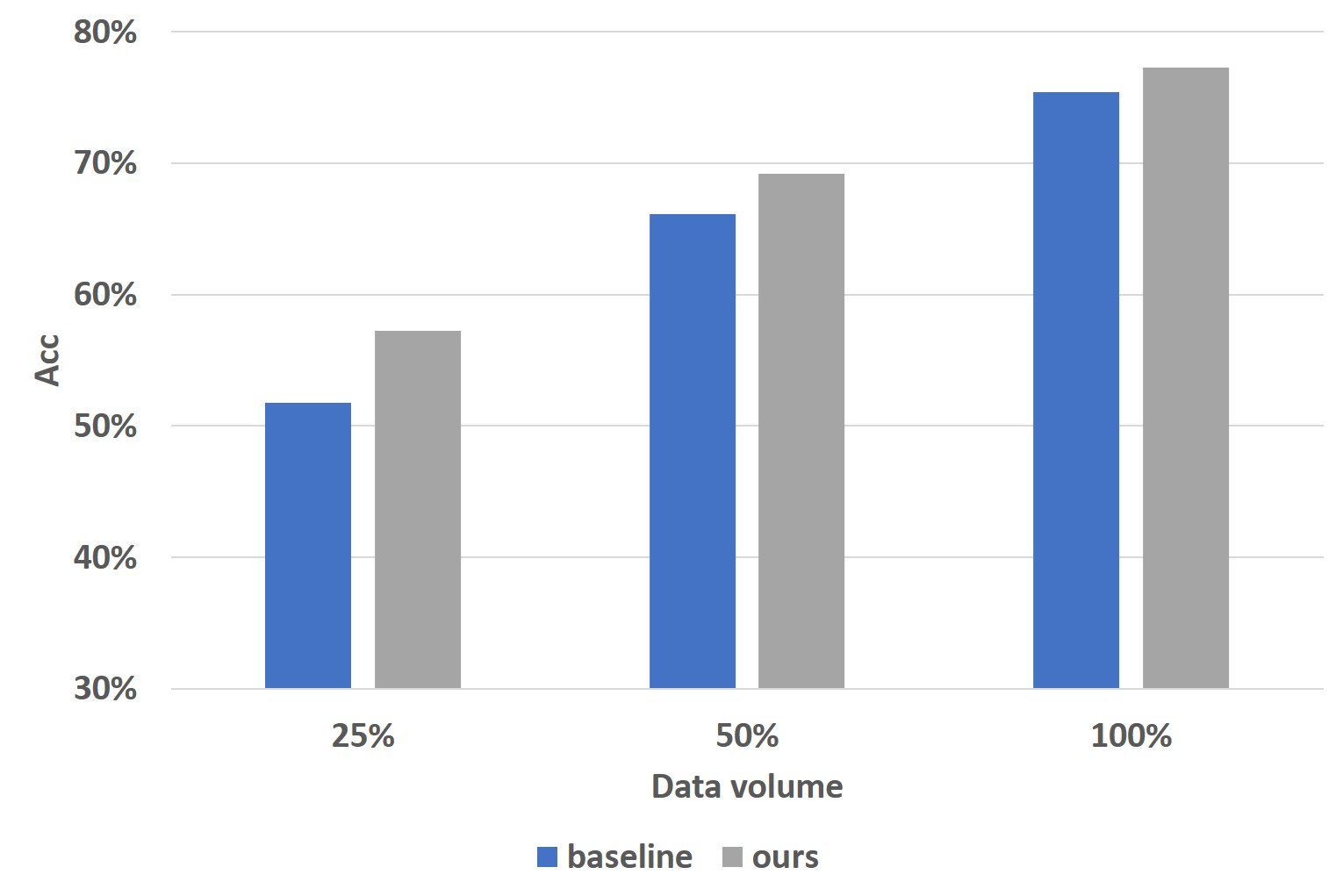}
  \caption{Experiments with ResNet18 on a small amount of Cifar100 data.}
  \label{fig:cifar_small_resnet18}
\end{figure}

In practical situations, acquiring extensive training data often poses challenges. As a result, we conduct experiments to explore the efficacy of cross-modal knowledge transfer in scenarios involving limited data. In Fig. \ref{fig:cifar_small_resnet18}, we compare the results of training with 25\% (using 25\% of the images per category), 50\% and 100\% of the Cifar100 dataset data, respectively. The results clearly demonstrate that our method exhibits significant improvement across varying amounts of training data. Notably, the performance improvement becomes more substantial as the amount of training data decreases.

We further investigate whether the model is capable of acquiring modality-share features during the distillation process. We assess the model's performance in matching different categories of text for each image in the test set of Cifar100. The results are presented in Table \ref{tab:cross_modal_match_cifar}. Considering that there are a total of 100 categories, a random selection will result in an accuracy of only $1\%$. It is evident that our method achieves a remarkably high cross-modal matching accuracy, which closely resembles the classification effect of the classifier. This outcome demonstrates the model's ability to effectively learn the correlation between image and text modalities. In contrast, certain existing knowledge distillation methods hardly learn this correlation because of fitting difficulties. A comparatively good cross-modal matching accuracy can be achieved using a contrast loss approach such as CLIP, but our approach achieves similar or better results, and the better performance of our approach on the task itself (Table \ref{tab:cifar}) illustrates the simultaneous learning of both modality-sharing features and modality-specific features.

\begin{table}[ht]
    \centering
    \caption{Accuracy of cross-modal matching for the ResNet34 model on the Cifar100 dataset.}
    \begin{tabular}{ccc}
    \hline
    method & Acc \\
    \hline
    chance & 1.00\%  \\
    FitNet \cite{2015FitNets} & 3.86\% \\
    ICKD \cite{liu2021exploring} & 1.04\% \\
    PKD \cite{passalis2018probabilistic} & 1.10\% \\
    CCL \cite{chen2021distilling} & 73.76\% \\
    CLIP \cite{radford2021learning} & 75.38\% \\
    \hline
    Ours (features) & 74.59\% \\
    Ours (classifier) & \textbf{78.50\%} \\
    \hline
    \end{tabular}
    \label{tab:cross_modal_match_cifar}
\end{table}

Subsequently, we proceed to carry out extensive experiments on a larger-scale Tiny-Imagenet dataset to corroborate the efficacy of our approach. The results are shown in Table \ref{tab:tiny_imagenet_resnet34}. The conclusions are similar to the previous ones in that our method achieves a significant performance improvement, whereas other existing methods fail to improve or have a limited improvement.

\begin{table}[ht]
    \centering
    \caption{Experiments with ResNet34 on Tiny-Imagenet dataset}
    \begin{tabular}{cc}
    \hline
    method & Acc \\
    \hline
    ResNet34 (w/o dist.) & 64.95\% \\
    \hline
    FitNet \cite{2015FitNets} & 65.52\% \\
    ICKD \cite{liu2021exploring} & 65.07\% \\
    PKD \cite{passalis2018probabilistic} & 64.07\% \\
    CLIP \cite{radford2021learning} & 65.10\% \\
    \hline
    ours (feature) & 65.98\% \\
    ours (feature+quality) & 66.15\% \\
    ours (classifier) & \textbf{66.48\%} \\
    \hline
    \end{tabular}
    \label{tab:tiny_imagenet_resnet34}
\end{table}

\subsubsection{Abation Study}
We explore the effect of hyperparameters on performance and the results are shown in Table \ref{tab:proj_cifar}, \ref{tab:l2_cifar} and \ref{tab:margin_cifar}. In terms of the mixing ratios of the projection heads, the effect is similar to the previous task, with an intermediate mixing ratio achieving the best results; too small a ratio will lead to fitting difficulties, while too large a ratio will affect the performance of the teacher model. For the weight of l2 loss in classifier level distillation, adding this item can further enhance the effect due to the strong correlation between the two modalities. For the value of margin, it can be seen that neither too large nor too small a margin value gives the best results; too small a margin can lead to fitting difficulties and overfitting problems, while too large a margin can lead to constraints that are too loose to learn anything useful. An intermediate value can have the best results. Also, we find that larger models can have smaller margin values, probably because of their better fitting ability.

\begin{table}[ht]
    \centering
    \caption{Effect of mixing ratio of projection head using ResNet34.}
    \begin{tabular}{ccc}
    \hline
    $\alpha$ & Acc \\
    \hline
    0.2 & 78.08\%  \\
    0.4 & 78.18\%  \\
    0.5 & 78.28\%  \\
    0.6 & \textbf{78.33\%}  \\
    0.8 & 77.93\% \\
    \hline
    \end{tabular}
    \label{tab:proj_cifar}
\end{table}

\begin{table}[ht]
    \centering
    \caption{Effect of weight of l2 loss using ResNet34.}
    \begin{tabular}{ccc}
    \hline
    $\beta$ & Acc \\
    \hline
    0 & 78.33\%  \\
    1 & \textbf{79.14\%}  \\
    2 & 78.82\%  \\
    \hline
    \end{tabular}
    \label{tab:l2_cifar}
\end{table}

\begin{table}[ht]
    \centering
    \caption{Effect of margin values.}
    \begin{tabular}{ccc}
    \hline
    $m$ & Acc (ResNet18) & Acc (ResNet34) \\
    \hline
    $0$ & 75.62\% & 77.96\% \\
    $0.1^2$ & 75.57\% & 78.26\% \\
    $0.2^2$ & 75.99\% & \textbf{78.33\%} \\
    $0.3^2$ & \textbf{76.28\%} & 77.77\% \\
    $0.4^2$ & 75.84\% & 77.66\% \\
    \hline
    \end{tabular}
    \label{tab:margin_cifar}
\end{table}

\section{Conclusion}
In this paper, we try to tackle the problem of migrating knowledge between widely differing modalities, which suffers from a variety of difficulties such as large modal differences, the tendency to overfit irrelevant features, and unaligned input data quality. To this end, we propose a framework for cross-modal knowledge transfer. We set up a learnable projection head to extract modal-related features from the teacher model. We propose two soft constrained knowledge distillations, both at the feature level and at the classifier level, which are capable of learning different modal knowledge while reducing fitting difficulty and avoiding overfitting irrelevant features. In addition, we also designed a quality-based adaptive weighting module, which can adaptively adjust the training objectives according to the quality of the input data, and solve the interference due to the misalignment of the quality of the input data.

We test knowledge transfer between the commonly used and widely varying modalities of image, speech, and text on speaker recognition and image classification tasks, achieving good performance gains in both cases. Existing knowledge distillation methods, on the other hand, have struggled to achieve performance improvements, suggesting that soft constraints are more effective between modalities with large variances. Our approach not only improves the performance of the original task but also enables cross-modal matching, which also suggests that modality-share features are learned during the distillation process. We also find some interesting findings such as the teacher model does not necessarily outperform the student model in cross-modal scenarios because there is a great deal of complementarity between the different modalities. We hope our cross-modality knowledge transfer strategies can introduce insights into other multi-modal learning tasks.

{\small
\bibliographystyle{IEEEtran}
\bibliography{ref}
}

\vfill


\end{document}